\documentclass{IET}

\usepackage{hyperref}
\usepackage{graphicx}
\usepackage{times}
\usepackage{epsfig}
\usepackage{amsmath}
\usepackage{amssymb}
\usepackage{marvosym}
\usepackage{framed}
\usepackage{mdframed}
\usepackage{subfigure}
\usepackage{epstopdf}
\usepackage[space]{grffile}
\usepackage[T1]{fontenc}
\usepackage{bbold}
\usepackage[export]{adjustbox}
\usepackage[utf8]{inputenc}
\usepackage[numbers]{natbib}
\DeclareGraphicsExtensions{.eps}
\begin{document}

\title{Bi-objective Framework for Sensor Fusion in RGB-D Multi-View Systems: Applications in Calibration}

\author[1,*]{Hassan Afzal}
\author[1]{Djamila Aouada}
\author[1]{Michel Antunes}
\author[2]{David Fofi}
\author[3]{Bruno Mirbach}
\author[1]{Bj\"{o}rn Ottersten}
\affil[1]{Interdisciplinary Centre for Security, Reliability and Trust University of Luxembourg, 4, rue Alphonse Weicker, L-2721, Luxembourg}
\affil[2]{Le2i - IUT Le Creusot, Université de Bourgogne,12 rue de la Fonderie, 71200, France}
\affil[3]{IEE S.A., 11, rue Edmond Reuter, L-5326 Contern, Luxembourg}
\affil[*]{Email: hassan.afzal@uni.lu}

\abstract{Complete and textured 3D reconstruction of dynamic scenes has been facilitated by mapped RGB and depth information acquired by RGB-D cameras based multi-view systems. One of the most critical steps in such multi-view systems is to determine the relative poses of all cameras via a process known as extrinsic calibration. In this work, we propose a sensor fusion framework based on a weighted bi-objective optimization for refinement of extrinsic calibration tailored for RGB-D multi-view systems. The weighted bi-objective cost function, which makes use of 2D information from RGB images and 3D information from depth images, is analytically derived via the Maximum Likelihood (ML) method. The weighting factor appears as a function of noise in 2D and 3D measurements and takes into account the affect of residual errors on the optimization. We propose an iterative scheme to estimate noise variances in 2D and 3D measurements, for simultaneously computing the weighting factor together with the camera poses. An extensive quantitative and qualitative evaluation of the proposed approach shows improved calibration performance as compared to refinement schemes which use only 2D or 3D measurement information.}

\maketitle

\section{Introduction}
RGB-D cameras provide simultaneous image and range data of the environment, offering enhanced sensing capabilities when compared to using single sensor modality. A non-exhaustive list of applications includes 3D telepresence systems \cite{Fuchs11}, creation of viewpoint free 3D videos \cite{Kuster11}, simultaneous localization and mapping \cite{Henry12}, or the acquisition of textured 3D surface models of static and dynamic scenes \cite{KinectFusion11,OmniKinect12,CopyMe13,Fuchs14}.

The acquisition of complete and textured 3D models of scenes required in domains such as security and surveillance, health, and entertainment, can be accomplished by using two different approaches. The first consists of using a single moving RGB-D camera with its location constantly being tracked \cite{Henry12,KinectFusion11,CopyMe13}. This solution is simple and attractive, however, has the drawback of not allowing to fully reconstruct dynamic scenes at each time-step. As an example, Dou et al. \cite{Dou_2015_CVPR} presented a 3D scanning system for deformable objects using a single Kinect sensor. The scanning results are promising, remark that, however, the obtained 3D reconstructions are rigid, even if the objects in the environment were constantly moving and deforming. This issue can be solved by using multiple fixed RGB-D cameras covering the entire scene \cite{OmniKinect12,Fuchs11,Fuchs14,Palasek15}. In this case, the relative poses of all RGB-D cameras are required for aligning the partial 3D reconstructions. The problem of estimating the relative poses of cameras in a multi-view system is known as extrinsic calibration.

Most of the works for extrinsic calibration of RGB-D multi-view systems rely on well established 2D camera based calibration routines~\cite{Zhang99, Bougouet} and pose refinement procedures, e.g. Bundle Adjustment (BA) \cite{BA, Amplianitis14, Fuchs14}, using 2D feature points extracted from the RGB images \cite{Fuchs11, Berger11, BergerICCV11}. The 3D information from the depth sensor has mainly been used in subsequent refinement steps using, e.g., the Iterative Closest Point (ICP) algorithm \cite{genicp,Penelle11,Yang13}. In this regard, the following question arises: how to optimally use both sources of complementary information.

In this paper, we investigate a strategy for RGB-D sensor fusion for the extrinsic calibration of multiple cameras. Instead of using 2D data from RGB images and 3D data from depth images independently, we propose a weighted bi-objective optimization scheme. We analytically derive a Least Squares (LS) based cost function, via the Maximum Likelihood (ML) method, that optimally combines the BA based 2D cost function with the ICP based 3D cost function. The sensor fusion is achieved by using a weighting factor that depends on two types of noise, one contaminating the 2D feature locations in the RGB images, and the second one contaminating the 3D point positions provided by the depth sensor. The experiments suggest that using the proposed joint cost for relative pose refinement provides more accurate results than the refinement schemes using 2D and 3D information separately. 

In the absence of information regarding noise levels in the 2D and 3D feature points we propose an iterative scheme which simultaneously estimates the noise along with the estimation of calibration parameters. The proposed scheme is completely automated requiring no manual intervention and no heuristic parameter setting. The quantitative and qualitative experiments show that the proposed scheme is able to perform sensor fusion for accurate camera calibration without any prior information about noise characteristics.

The present work extends and consolidates our previous work called \textit{BAICP+}~\cite{BAICP14}, which experimentally showed that in many cases, using a heuristically constructed weighted bi-objective refinement approach that combines 2D and 3D information provides better results than refinement approaches based on cost functions using only 2D or 3D information.
\\
\subsection{Contributions}
\begin{itemize}
	\item We present a sensor fusion framework based on weighted bi-objective optimization for refinement of extrinsic calibration of an RGB-D multi-view system. We derive an analytic expression for the weighting factor, in the bi-objective optimization, in terms of noise in measurements of RGB and depth sensors. 
	\item We propose an iterative scheme for extrinsic calibration of an RGB-D multi-view system, which alternates between camera pose estimation, and the computation of measurement noise levels.
	\item We perform a thorough experimental evaluation on synthetic and real data, and show that fusing the RGB-D information using the proposed bi-objective optimization provides superior results when compared to refinement schemes that only use 2D or 3D feature information.
\end{itemize}
\subsection{Article Overview}
We start by giving a brief overview of state-of-the-art methods for extrinsic calibration used in RGB-D multi-view systems, as well as of bi-objective pose estimation approaches in Section~\ref{Related Work}. In Section~\ref{Problem formulation}, the extrinsic calibration problem is formally presented. Section~\ref{Background} gives a brief introduction of BA and ICP together with our previous work i.e., \textit{BAICP+}~\cite{BAICP14}, in which BA and ICP are heuristically combined. Section~\ref{Proposed Approach} analytically derives the expression for the weighted bi-objective cost function for refinement of extrinsic calibration parameters. Section~\ref{AutomaticWeighting} presents an automated iterative approach for camera pose estimation and estimation of the measurement noise parameters. In Section~\ref{Experiments with synthetic data} and Section~\ref{Experiments with real data}, we analyze and illustrate the benefits of the proposed approach via extensive experiments using synthetic and real data respectively. This is followed by a conclusion in Section~\ref{Conclusion}.

\begin{figure*}[t]
	\centering
	\subfigure[ RGB-D Multi-View System (4 cameras)]{\includegraphics[width=0.44\columnwidth]{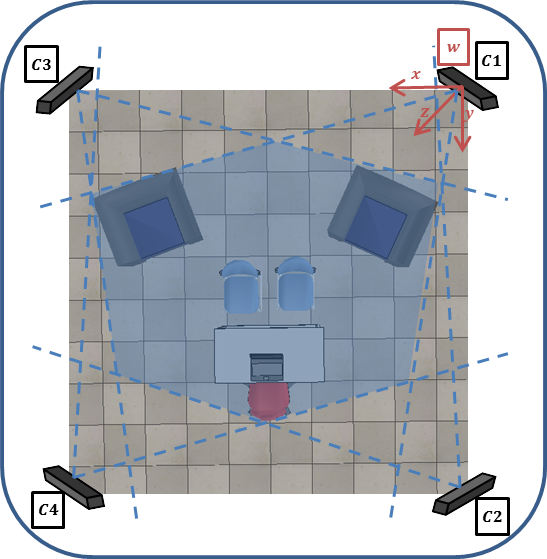}}~~~~~
	\subfigure[Full 3D Scene Reconstruction]{\includegraphics[width=0.5\columnwidth]{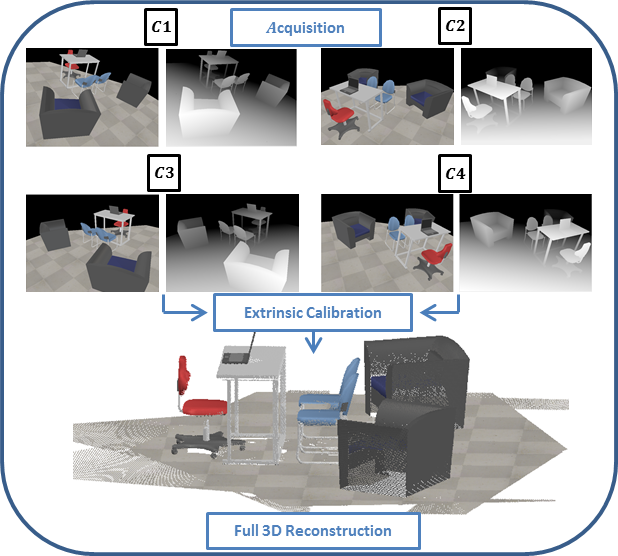}}
	\caption{RGB-D Multi-View System with full scene 3D reconstruction in a simulated setup. (a) RGB-D Multi-View System (4 cameras)  with field of view (FOV) of each camera. The highlighted region represents overlapping FOVs of all cameras. The global reference frame $\mathtt{w}$ is aligned with camera \textbf{\textit{C}1}. (b) Steps required for Full 3D Scene Reconstruction using an RGB-D Multi-View System. Each camera acquires a RGB image and a depth image, which are used to estimate the relative pose of each camera with respect to $\mathtt{w}$. After extrinsic calibration, estimated poses are used to put all acquisitions in $\mathtt{w}$ to get complete reconstruction.}
	\label{fig_C1} 
\end{figure*}
\section{Related Work}\label{Related Work}
In this section we review state-of-the-art techniques for extrinsic calibration in an RGB-D multi-view system with a focus on the modality of data used. We also briefly overview the sensor fusion approaches, based on bi-objective optimization, for solving the pair-wise pose estimation problem.

A considerable amount of research has been, and is still being, carried out in the domain of RGB-D cameras based multi-view systems~\cite{Berger14}. We are interested in analyzing the extrinsic calibration method used in such multi-view systems.
Extrinsic calibration in a multi-view system requires information about the same points (feature points), in 3D space, to be acquired in different views. This information can be extracted from 2D or 3D camera acquisitions of objects with, e.g., known textural and/or geometrical properties~\cite{Berger11, OmniKinect12, Alexiadis12, Miller13, Palasek15}. A planar checkerboard pattern first proposed by Zhang~\cite{Zhang99} is one of the most widely used objects for this purpose. Assuming correct mapping of RGB and depth sensors, Maimone and Fuchs~\cite{Fuchs11}, and Yang et al.~\cite{Yang13} make use of such an object, with corners extracted from RGB and IR images respectively, to calibrate their systems with the help of Bougouet's calibration toolbox~\cite{Bougouet}. Berger et al.~\cite{BergerICCV11,Berger11}, on the other hand, try to improve accuracy by calibrating all RGB and depth sensors together by extracting 2D feature points from both RGB and IR images with the help of a  special checkerboard pattern consisting of diffuse and mirroring patches.

A major drawback of 2D only calibration approaches is their inability to tackle noise specific to depth sensors which causes problems in alignment of 3D data from multiple cameras. Various methods try to tackle this problem with the help of an explicit depth correction step based on comparing known and measured depths for each camera, e.g., ~\cite{Fuchs11, OmniKinect12, Nakazawa12, Fuchs14} etc. There are other methods, such as~\cite{Yang13}, which add a final refinement step based on the ICP algorithm, using 3D data only, which tries to mitigate the pose misalignment due to depth specific sensor noise. Penelle et al.~\cite{Penelle11} propose to use only 3D points, corresponding to 2D feature points (from checkerboard) extracted from RGB images, in the ICP based calibration scheme where initial estimates are provided via the RANSAC algorithm. Nakazawa et al.~\cite{Nakazawa12}, on the other hand, use 2D feature points in the BA algorithm for pose refinement but perform corrections using alignment of 3D data as well. Miller et al.~\cite{Miller13} propose to use 3D information related to foreground objects to obtain an occlusion aware energy minimization scheme for auto-calibration. Deng et al.~\cite{Deng2014}, use 3D feature points observed at different locations in the scene to compute a smooth field of rigid transformation instead of a single rigid transformation to align the measurements of two RGB-D cameras. 

Dou and Fuchs ~\cite{Fuchs14}, in their work on multi-view 3D reconstruction, proposed to combine 2D and 3D information in a weighted bi-objective optimization scheme derived from their previous work on pair-wise pose tracking for mono-view 3D reconstruction~\cite{Dou2012}. They propose to use matching feature points extracted via Scale-Invariant Feature Transform (SIFT) from RGB images with matching planes extracted from 3D/depth images in a weighted bi-objective BA scheme. The weighting factor is selected empirically for all experiments. A similar approach is proposed by Henry et al.~\cite{Henry12}, using a global ICP scheme to align 2D visual feature points and 3D/depth measurements from multiple views but the weights are, again, selected empirically. Tykkala et al.~\cite{Tykkala} use what they call an image based direct ICP approach for pairwise pose estimation. They propose to compute the weighting factor via a heuristic measure using ratio of the median intensity and the depth values of selected points. Michot et al.~\cite{Michot} propose to use a weighted bi-objective BA scheme for the multi-sensor Simultaneous Localization and Matching (SLAM) problem. They discuss the dependence of the weighting factor on the ratio of the noise variance for each sensor's measurement and formulate their bi-objective optimization by using a Mean Squared Error (MSE) based cost function from individual sensors. They investigate three methods for automatic weight computation namely L-Curve, L-Tangent Norm and cross validation with experiments showing that the L-Curve based method performs better than the others.

This brief survey shows that most work done on extrinsic calibration of RGB-D multi-view systems is based on calibration schemes which use 2D or 3D information independently. These methods do not take into account the relative accuracy of 2D and 3D measurement in a systematic manner to, e.g., give more importance to less noisy measurements. 

In this work, we perform sensor fusion by formulating a weighted bi-objective optimization scheme based on 2D and 3D cost functions for performing global refinement of poses in RGB-D multi-view systems. We draw upon work in the Robotics domain~\cite{Michot, Tykkala, Han, Dou2012}, where several sensor modalities are exploited for pair-wise pose estimation. The proposed bi-objective optimization uses cost functions from both the BA and ICP algorithms, based on 2D and 3D measurements, in a single unified cost function. The key to combining the two cost functions is the information about noise in the 2D and 3D measurements which is reflected in the weighting factor. In the absence of this information we propose to use a simple approach which iteratively estimates the noise parameters given the current camera poses and vice versa. We show the validity of the proposed approach by achieving improved results in the experiments performed under different conditions.

\section{Extrinsic Calibration Problem of Multiple RGB-D Sensors}\label{Problem formulation}
\textit{Notation}: The following notation will be adopted. Subscripts indicate camera or reference frame indexes and superscripts indicate point indexes:
\begin{center}
    \begin{tabular}{ l  l }
	$\mathtt{w}$ : &world reference frame.\\
	$\mathbf{A}$ : & matrix.\\
	$\mathbf{p}$  : & vector.\\
	$l, M$ : &scalars.\\
	$tr(\mathbf{A})$ : & trace of matrix $\mathbf{A}$.\\
	$\mathbf{A}^\intercal$ : & transpose of matrix $\mathbf{A}$.\\
	$\psi(\cdot)$, :  & $\mathtt{w}$ to image plane projection.\\
	$\hat{\mathbf{p}}$, $\hat{\mathtt{\mathbf{T}}}_l$:  & estimates of $\mathbf{p}$ and $\mathtt{\mathbf{T}}_l$, respectively.\\
	$\mathbf{I}_n$ : & identity matrix of dimension $n \times n$.\\
	$\mathbf{0}_n$ : & null vector of dimension $n \times 1$.
	\end{tabular}	
\end{center}

In this section, we will formulate the extrinsic calibration problem for an RGB-D multi-view system. Let us consider a multi-view system composed of $N$, intrinsically calibrated,  RGB-D cameras with intersecting FOVs, as shown in \figurename~\ref{fig_C1}. Every RGB-D camera $l$, with $l=1, \cdots, N$, acquires an RGB image $\mathbf{C}_{l}$ and a 3D vertex map $\mathbf{V}_{l}$, with associated known matrix of intrinsic parameters $\mathbf{K}_l$.

In order to correctly align the partial 3D reconstructions $\{\mathbf{V}_l\}$, where $l=1, \cdots, N$,  acquired by $N$ RGB-D cameras, it is necessary to accurately estimate their positions with respect to a global reference frame, referred to as $\mathtt{world}$ and denoted by $\mathtt{w}$, as shown in \figurename~\ref{fig_C1}. Each camera's relative position with respect to $\mathtt{w}$ is defined by:
\begin{align}\label{eq2.0}
{\mathbf{T}}_{l}  = \begin{pmatrix} 
\mathbf{R}_l& \mathbf{t}_{l}\\ \mathbf{0}^\intercal_3&1 
\end{pmatrix},
\end{align}
where $\mathbf{T}_{l} \in SE(3)$ represents the rigid transformation, from camera $l$ to $\mathtt{w}$. The matrix $\mathbf{R}_l$ is rotation matrix in $SO(3)$ and $\mathbf{t}_{l} \in \mathbb{R}^3$ is translation vector. Therefore the same point $\mathbf{p} \in \mathbb{R}^3$ in $\mathtt{w}$ viewed by camera $l$ as $\mathbf{p}_{l}$ and by cameras $k$ as $\mathbf{p}_{k}$ can be related to the cameras' reference frames as follows:
\begin{equation}\label{eq2.1}
\mathbf{R}_l\mathbf{p}_{l} + \mathbf{t}_{l}=\mathbf{R}_k \mathbf{p}_{k}+ \mathbf{t}_{k}.
\end{equation}
Similarly, for a given point $\mathbf{x} \in \mathbb{R}^3$ in $\mathtt{w}$,  its projection on each camera's image plane results in 2D pixel coordinates $\mathbf{q}_{l}$, such that:
\begin{equation}\label{eq2.2}
\mathbf{q}_{l} = \psi\left(\mathbf{K}_{l}, \mathbf{T}_{l}, \mathbf{x}\right),\;\;\forall l,
\end{equation}
where $\psi(.)$ is $\mathtt{world}$ to image plane projection function.

Let us assume that all cameras are of resolution $M$. The color pixel positions in $\mathbf{C}_{l}$ may be represented by the points $\mathbf{q}_{l}^{i} \in [\mathbf{q}_{l}^{1},\cdots,\mathbf{q}_{l}^{M}]$ where $i\in\{1,\cdots,M\}$. Similarly, the 3D coordinates in $\mathbf{V}_{l}$ may be represented by points $\mathbf{p}_{l}^{k} \in [\mathbf{p}_{l}^{1},\cdots,\mathbf{p}_{l}^{M}]$ where $k\in\{1,\cdots,M\}$.

The problem at hand may therefore be stated as follows. Given $N$ RGB-D  cameras in a multi-view system with acquired RGB images $\{\mathbf{C}_{1}, \cdots, \mathbf{C}_{N}\}$ and 3D vertex maps $\{\mathbf{V}_{1}, \cdots, \mathbf{V}_{N}\}$,
we assume knowledge of $H \le M$ matching points in each camera's RGB image plane referred to as 2D feature points and denoted as $[\mathbf{q}^{1}_{l}, \cdots, \mathbf{q}^{H}_{l}]$. Similarly, we assume knowledge of $J \le M$  matching 3D points in each camera's 3D vertex map called 3D feature points and denoted as $[\mathbf{p}^{1}_{l}, \cdots, \mathbf{p}^{J}_{l}]$. Moreover we assume knowledge of each camera's intrinsic parameters, ${\mathbf{K}}=[{\mathbf{K}}_{1},\cdots, {\mathbf{K}}_{N}]$.
Using this information, we want to find the estimates of the parameters ${\mathbf{\mathbf{T}}} = [{\mathbf{\mathbf{T}}}_{1}, \cdots, {\mathbf{\mathbf{T}}}_{N}]$.
\section{Background}\label{Background}
In this section, we introduce the two pose refinement algorithms namely BA and ICP, from which we derive the 2D and 3D cost functions used in our proposed bi-objective cost function.  We also introduce \textit{BAICP+}~\cite{BAICP14}, in which we proposed a bi-objective pose refinement scheme, but, with heuristically defined weight and scaling parameters. 
\subsection{Bundle Adjustment (BA)}
For refinement of extrinsic calibration parameters using 2D feature points only, we use a cost function from the Bundle Adjustment (BA) algorithm~\cite{BA}. It  has been the method of choice for problems related to multi-view 3D reconstruction and pose refinement based on 2D feature points extracted from RGB images~\cite{BA}.
Bundle Adjustment (BA) requires an initial estimate of the pose parameters. Moreover, it also requires an estimate of 3D points i.e., $[\mathbf{x}^{1},\cdots,\mathbf{x}^{H}]$, corresponding to available 2D feature points $[\mathbf{q}^{1}_{l}, \cdots, \mathbf{q}^{H}_{l}]$. These estimates are then refined by computing the error of projection of estimate of each 3D point $\mathbf{x}^{h}$, $h=1, \cdots, H$, corresponding to the 2D feature point $\mathbf{q}_{l}^{h}$ to camera $l$ via:
\begin{equation}\label{eq2.3}
\mathbf{a}_{l}^{h}(\mathbf{S}_{l}^{h})=\mathbf{q}_{l}^{h} - \psi\left(\mathbf{K}_{l}, \mathbf{T}_{l}, \mathbf{x}^{h}\right),
\end{equation}
where $\mathbf{a}_{l}^{h}(\mathbf{S}_{l}^{h}) \in \mathbb{R}^2$ and $\mathbf{S}_{l}^{h}=\left(\mathbf{\mathbf{T}}_{l}, \mathbf{x}^{h}\right)$ \footnote{BA can also refine the estimate of intrinsics $\mathbf{K}_{l}$ if required}. Therefore, the total BA cost to be minimized for the refinement of estimates of each camera's pose parameters together with the estimates of 3D points corresponding to 2D feature points is given as:
\begin{equation}\label{eq2.4}
V_{BA}({\mathbf{S}}) = \sum_{l=1}^{N}tr(\mathbf{A}_{l}^\intercal({\mathbf{S}}_{l})\mathbf{A}_{l}({\mathbf{S}}_{l})),
\end{equation}
where $\mathbf{S}=\left(\mathbf{\mathbf{T}}, \mathbf{X}\right)$, $\mathbf{S}_{l}=\left(\mathbf{\mathbf{T}}_{l}, \mathbf{X}\right)$, $\mathbf{X}=[\mathbf{x}^{1},\cdots,\mathbf{x}^{H}]$ and $\mathbf{A}_{l}(\mathbf{S}_{l})=[\mathbf{a}_{l}^{1}(\mathbf{S}_{l}^{1}),$ $\cdots, \mathbf{a}_{l}^{H}(\mathbf{S}_{l}^{H})]$.
\subsection{Iterative Closest Point (ICP)}
For refinement of the extrinsic calibration parameters using 3D feature points only, we use the cost function from the global Iterative Closest Points (ICP) algorithm~\cite{ICP92}. ICP algorithm has been the de facto solution for pose refinement problems when only 3D feature points are available~\cite{ICP, ICP92}. ICP algorithm also uses initial estimates of the pose parameters and minimizes the Euclidean distance between corresponding 3D feature points from different views, such that:
\begin{equation}\label{eq2.5}
\mathbf{b}_{l,k}^{j}(\mathbf{T}_{l},\mathbf{T}_{k})= (\mathbf{R}_l\mathbf{p}^j_{l} + \mathbf{t}_{l})-(\mathbf{R}_k \mathbf{p}^j_{k}+\mathbf{t}_{k}),
\end{equation}
where $\mathbf{b}_{l,k}^{j}(\mathbf{T}_{l},\mathbf{T}_{k}) \in \mathbb{R}^3$ and $j\in[1,\cdots,J]$. Therefore, the total ICP cost to be minimized for refinement of each camera's pose parameters is given as:
\begin{align}\label{eq2.6}
V_{ICP}({\mathbf{T}}) = \sum_{\substack{1<l,k<N \\l\neq k}}tr(\mathbf{B}_{l,k}^\intercal({\mathbf{\mathbf{T}}}_{l},{\mathbf{\mathbf{T}}}_{k})\mathbf{B}_{l,k}({\mathbf{\mathbf{T}}}_{l},{\mathbf{\mathbf{T}}}_{k})),
\end{align}
where $\mathbf{B}_{l,k}({\mathbf{\mathbf{T}}}_{l},{\mathbf{\mathbf{T}}}_{k})=[\mathbf{b}_{l,k}^{1}({\mathbf{\mathbf{T}}}_{l},{\mathbf{\mathbf{T}}}_{k}), \cdots, \mathbf{b}_{l,k}^{J}({\mathbf{\mathbf{T}}}_{l},{\mathbf{\mathbf{T}}}_{k})]$.
\subsection{BAICP+}
In our previous work~\cite{BAICP14}, we used the information provided by RGB-D cameras in a bi-objective optimization scheme for extrinsic calibration refinement. It combines the BA and ICP cost functions based on 2D and 3D feature points respectively resulting in the cost function for \textit{BAICP+}:
\begin{align}\label{eq12.1}
V_{BAICP}({\mathbf{S}}) = \frac{(1-c)}{a}V_{ICP}({\mathbf{T}})+ \frac{sc}{b}V_{BA}({\mathbf{S}}),
\end{align}
where $c \in [0,1]$ is the weighting factor and $s=(\frac{avgDepth}{avgFocal})^2$ is a heuristic scaling factor, based on the ratio of the average depth of points in ${\mathbf{x}}$ versus average focal length across all views. The parameters $a$ and $b$ denote the total number of 3D point correspondences and 2D feature points points across all views.
\section{Bi-Objective Extrinsic Calibration}\label{Proposed Approach}
In this section, we present the bi-objective optimization for refinement of the extrinsic calibration parameters in an RGB-D multi-view system. We use cost functions defined in the previous section which use 2D and 3D feature points extracted from RGB images and vertex maps, respectively. 

In this work, we propose to formally analyze and derive an expression for the cost function, based on ML estimations, of the bi-objective optimization taking into account the noise affecting both 2D and 3D measurement/feature points.
We assume the presence of independent additive Gaussian noise in each coordinate of the 3D feature points such that:
\begin{equation}\label{eq4.1}
\tilde{\mathbf{p}}_{l}^{j}\sim\mathcal{N}\left({\mathbf{p}}_{l}^{j},\sigma^2_{3D}\mathbf{I}_{3}\right), 
\end{equation}
where $\tilde{\mathbf{p}}_{l}^{j}$ is the noisy 3D point and ${\mathbf{p}}_{l}^{j}$ is the noise free point. Similarly for 2D feature points we have:
\begin{equation}\label{eq4.2}
\tilde{\mathbf{q}}_{l}^{h}\sim\mathcal{N}({\mathbf{q}}_{l}^{h},\sigma^2_{2D}\mathbf{I}_{2}), 
\end{equation}
where $\tilde{\mathbf{q}}_{l}^{h}$ is the noisy 2D point and ${\mathbf{q}}_{l}^{h}$ is the noise free point. This means that we have to use the noisy 2D and 3D feature points to estimate the pose parameters. This leads to redefining the 3D error function $\mathbf{b}_{l,m}^{j}(\mathbf{T}_{l},\mathbf{T}_{m})$, given in~\eqref{eq2.5}, such that it computes the error between noisy points $\tilde{\mathbf{p}}_l^j$ and $\tilde{\mathbf{p}}_l^m$ projected to $\mathtt{w}$, from camera $l$ and $m$, using the pose parameters $\mathbf{T}_l$ and $\mathbf{T}_m$, respectively. Similarly the 2D error function $\mathbf{a}_{l}^{h}(\mathbf{S}_{l}^{h})$, given in~\eqref{eq2.3} where $\mathbf{S}_{l}^{h}=(\mathbf{T}^l,\mathbf{x}^h)$, is redefined such that it computes the 2D error between back projection of the estimated 3D point $\mathbf{x}^h$ to camera $l$, using $\mathbf{T}_l$ and $\mathbf{K}_l$, and the corresponding noisy 2D feature point $\tilde{\mathbf{q}}_l^h$.

Now, we can define the distribution the 3D error  $\mathbf{b}_{l,m}^{j}(\mathbf{T}_{l},\mathbf{T}_{m})$ is drawn from by considering the noise free 3D points $\mathbf{p}_{l}^{j}$ and $\mathbf{p}_m^{j}$ in \eqref{eq2.1} such that~\cite{genicp}:
\begin{align}\label{eq4.3}
\mathbf{b}_{l,m}^{j}(\mathtt{T}_{l},\mathtt{T}_{m})&\sim\mathcal{N}\left((\mathbf{R}_l{\mathbf{p}}^j_{l}  +\mathbf{t}_{l})-(\mathbf{R}_m {\mathbf{p}}^j_{{m}}+\mathbf{t}_{m}),
\mathbf{R}_l\sigma^2_{3D}\mathbf{I}_{3}\mathbf{R}_l^\intercal+ 
\, \mathbf{R}_m\sigma^2_{3D}\mathbf{I}_{3}\mathbf{R}_m^\intercal\right) \nonumber \\
&=\mathcal{N}(\mathbf{0}_{3}, 2\sigma^2_{3D}\mathbf{I}_{3}).
\end{align}
Similarly, considering the noise free 2D measurements in \eqref{eq2.2}, we have $\mathbf{a}_{l}^{h}(\mathbf{S}_{l}^{h})\sim\mathcal{N}(\mathbf{0}_{2}, \sigma^2_{2D}\mathbf{I}_{2})$. It is clear from \eqref{eq4.3} that since $\mathbf{b}_{l,m}^{j}(\mathbf{T}_{l},\mathbf{T}_{m})$, which is based on the ICP algorithm, uses two noisy 3D feature points, hence, the variance of the corresponding distribution is two times the variance of noise in each 3D feature point. This is in contrast to the variance of distribution corresponding to $\mathbf{a}_{l}^{h}(\mathbf{S}_{l}^{h})$, which is based on the BA algorithm and uses only one noisy 2D feature point~\cite{Michot}.

Using $\mathbf{b}_{l,m}^{j}(\mathbf{T}_{l},\mathbf{T}_{m})$ and $\mathbf{a}_{l}^{h}(\mathbf{S}_{l}^{h})$, we want to find the likelihood cost function, maximum of which gives the Maximum Likelihood Estimate (MLE) of the parameters ${\mathbf{S}}=\left({\mathbf{T}}, {\mathbf{x}}\right)$. Since the MLE with Gaussian model is equivalent to the Least Squares Estimate (LSE)~\cite{CVModels}, we can directly get:
\begin{align}\label{eq4.4}
\hat{\mathbf{S}}=\arg\min_{{\mathbf{S}}}\sum_{\substack{1<l,m<N \\l\neq m}}&\frac{1}{2\sigma_{3D}^{2}}tr\left(\mathbf{B}_{l,m}^\intercal({\mathbf{T}}_{l},{\mathbf{T}}_{m})\mathbf{B}_{l,m}({\mathbf{T}}_{l},{\mathbf{T}}_{m})\right) 
+ \sum_{l=1}^{N} \frac{1}{\sigma_{2D}^{2}}tr\left(\mathbf{A}_{l}^\intercal({\mathbf{S}}_{l})\mathbf{A}_{l}({\mathbf{S}}_{l})\right).
\end{align}
Therefore, the total cost to be minimized is:
\begin{align}\label{eq4.5}
V({\mathbf{S}}) = V_{ICP}({\mathbf{T}})+ wV_{BA}({\mathbf{S}}),
\end{align}
where $w=\frac{2\sigma_{3D}^{2}}{\sigma_{2D}^{2}}$ is the weighting factor. The cost function in \eqref{eq4.4} optimally combines information from RGB and depth sensors, to be used in the pose refinement scheme, by taking into account the noise levels in the 2D and 3D points. It formally defines the the relationship of measurement noise in the 2D and 3D feature points with the weighting factor $w$. In case the assumption of noise with same variances affecting all 2D and 3D points respectively, does not hold and information about the  noise variances affecting each point is available, it can be incorporated in the proposed framework. Moreover, the use of the ICP based cost also allows the use of all the 3D points acquired by each sensor (with the help of nearest neighbor correspondence) in the optimization scheme when only 2D feature points are available.

The cost function~\eqref{eq4.5} is a non-linear function of the parameters $\mathbf{S}$ and we resort to numerical search methods~\cite{MLE} to optimize the criterion. Please refer to {Appendix}~\ref{Appendix} for further discussion.

\begin{figure}[th]
	\centering
	\subfigure{\includegraphics[width=0.44\columnwidth]{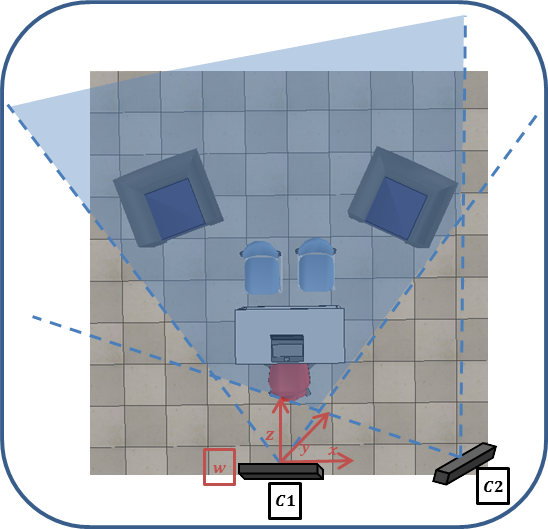}}
	\caption{RGB-D Multi-View System (2 cameras)  with field of view (FOV) of each camera. The highlighted region represents overlapping FOVs of all cameras. The global reference frame $\mathtt{w}$ is aligned with camera \textbf{\textit{C}1}. }
	\label{fig_C2} 
\end{figure}
\section{Weighting Factor Estimation}\label{AutomaticWeighting}
In this section we discuss the automatic and simultaneous estimation of the weighting factor $w$ in \eqref{eq4.5}, together with the camera poses in the absence of information regarding noise affecting both the 2D and 3D measurements. We propose an approach which alternates between camera pose estimation and estimation of the 2D and the 3D noise variances to arrive at a suitable solution.

In the previous section, the estimates of camera pose parameters and 3D points in $\mathtt{w}$ corresponding to 2D feature points, were computed based on known 2D and 3D feature points and the noise affecting them. We assumed the presence of Gaussian noise with zero mean and variances of $\sigma^2_{2D}$ and $\sigma^2_{3D}$ in 2D and 3D measurements, respectively. These parameters, in turn, define the weighting factor $w$ which is instrumental in constructing the sensor fusion framework by optimally combining the 2D and 3D cost functions to estimate the camera poses. In real-world scenarios, however, information about the noise affecting one or both sensor measurements is often unavailable. This makes the computation of a correct $w$ difficult. As mentioned in Section~\ref{Related Work}, researchers have tried to estimate the optimal weighting factor, for their proposed bi-objective schemes, for solving mainly the pair-wise pose estimation problem. The commonly used used methods range from using simple heuristic measures such as in the case of~\cite{Tykkala} to more complex methods, based on analysis of trade-off between residuals of two cost functions and based on learning via cross-validation, such as in the case of~\cite{Michot}.

In this work, we propose to use a simple method for automatic estimation of the weighting factor $w$ which finds its basis in finding the MLE of noise variances, $\sigma^2_{2D}$ and $\sigma^2_{3D}$, using the 2D and 3D feature points together with the current estimates of camera poses and 3D points in $\hat{\mathbf{S}}$. The MLE of  the variance $\sigma^2_{3D}$ is given as~\cite{CVModels}:
\begin{align}\label{eq3.04}
\hat{\sigma}^2_{3D}&= \sum_{\substack{1<l,m<N \\l\neq m}}\frac{tr(\mathbf{B}_{l,m}^\intercal({{\mathbf{T}}}_{l},{{\mathbf{T}}}_{m})\mathbf{B}_{l,m}({\mathbf{{T}}}_{l},{\mathbf{{T}}}_{m}))}{2a},
\end{align}
where $a$ is the total number of 3D feature correspondences across all views. Similarly, the MLE of the variance ${\sigma}^2_{2D}$ is computed via:
\begin{align}\label{eq3.05}
\hat{\sigma}^2_{2D}&= \sum_{l=1}^{N}\frac{tr(\mathbf{A}_{l}^\intercal({\mathbf{S}}_{l})\mathbf{A}_{l}({\mathbf{S}}_{l}))}{b}, 
\end{align}
where $b$ is the total number of 2D feature points found across all views.

We follow an iterative approach whereby using 2D and 3D feature points and an initial estimate $\hat{\mathbf{S}}$, the MLE estimates of noise variances and hence of $w$ are obtained via~\eqref{eq3.04} and~\eqref{eq3.05}. This initial estimate of $w$ is then used to find an updated estimate of $\mathbf{S}$ using~\eqref{eq4.5} via non-linear optimization which, in turn, is used to update the estimate of $w$. This process is repeated for a fixed number of iterations until the estimates of $\mathbf{S}$ and $w$ converge. 

\begin{figure*}[t]
	\centering
	\subfigure[Features extracted from RGB image]{\includegraphics[width=0.45\columnwidth]{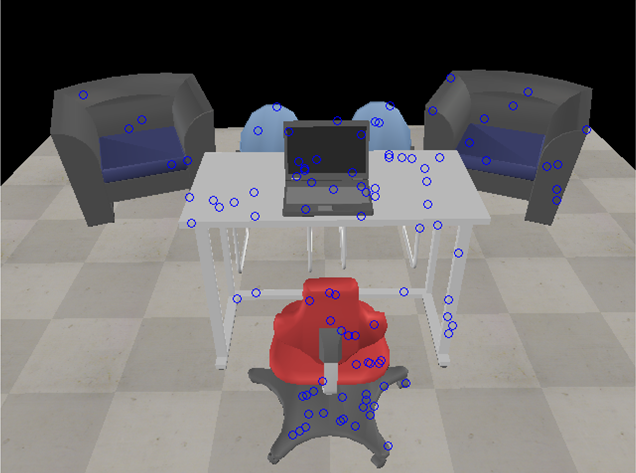}}~~
	\subfigure[Features extracted from depth image]{\includegraphics[width=0.45\columnwidth]{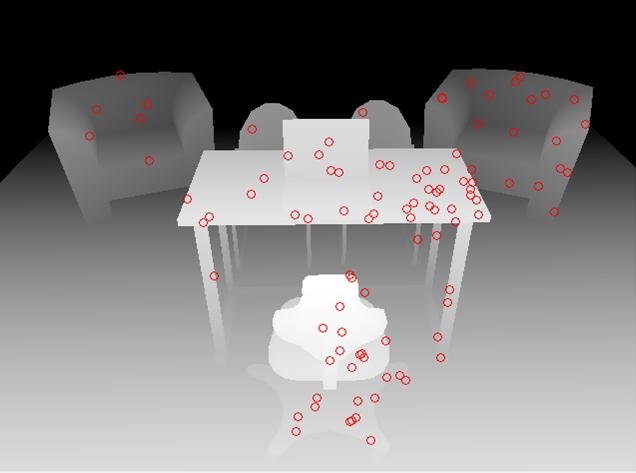}}
	\caption{Features extracted from RGB and depth images of camera \textbf{\textit{C}1} in the multi-view system composed of 2 cameras as shown in~\figurename~\ref{fig_C2}. The extracted feature points are also visible to camera \textbf{\textit{C}2}. }
	\label{fig_C3} 
\end{figure*}
\section{Experiments with Synthetic Data}\label{Experiments with synthetic data}
In this section, we carry out a quantitative performance analysis of the proposed bi-objective refinement with a known and an unknown weighting factor.
\subsection{Evaluation Methodology and Parameters}\label{Evaluation Methodology and Parameters}
We use V-REP~\cite{vrep} to simulate 2 and 4 cameras based RGB-D multi-view systems, with overlapping FOVs, as shown in \figurename~\ref{fig_C2} and \figurename~\ref{fig_C1}, respectively. In both cases, the global reference frame $\mathtt{w}$ lies in camera \textbf{\textit{C}1}. We simulate a scene containing several objects such as chairs, a table, sofas etc. The acquired noise-free data, in the form of RGB and depth images, is assumed to be perfectly mapped in each camera's RGB sensor's reference frame with known intrinsics. After data acquisition, random points, visible to all cameras, are extracted as feature points in both RGB and depth images as shown in \figurename~\ref{fig_C3} (points on the floor are discarded). Features extracted from depth maps are converted to the corresponding 3D points via known intrinsics. 

In the next step, noise is added to the extracted 2D and 3D feature points. We assume the presence of independent Gaussian noise in each coordinate of position of 2D feature points with zero mean and standard deviation $\sigma_{2D}$ similar to~\cite{Zhang04}. The value of $\sigma_{2D}$ is varied between $0.2$ to $1.8$ pixels  with a step size of $0.4$ pixels. Depth sensor measurements in RGB-D cameras suffer from different types of systematic and non-systematic errors as investigated in~\cite{3DKinect, 3DKinect2}. For our scheme we propose to counter, beforehand, the systematic errors in depth measurements of each camera via a correction step, based on comparing known and measured depths \cite{Fuchs11, Nakazawa12}. Therefore, for all remaining errors we assume the presence of additive independent Gaussian noise in each coordinate of 3D feature points in each view with zero mean and standard deviation $\sigma_{3D}$. The value of $\sigma_{3D}$ is varied between $6$ to $30$ mm with a step size of $6$ mm to keep it in the range of errors computed in~\cite{3DKinect2}. 
\begin{figure*}[t]
	\centering
	\subfigure[Rotation error]{\includegraphics[width=0.475\columnwidth]{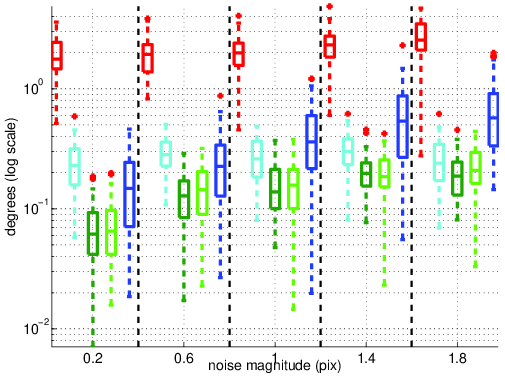}}
	\subfigure[Relative translation error]{\includegraphics[width=0.475\columnwidth]{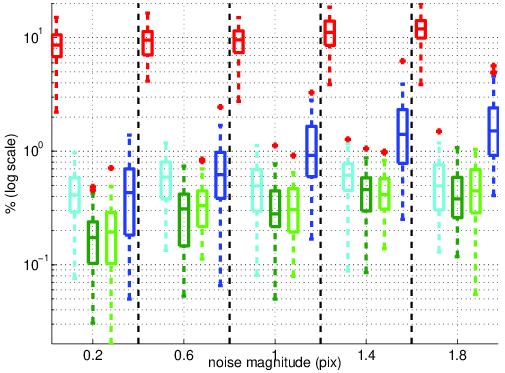}} \\
	\includegraphics[width=0.5\textwidth]{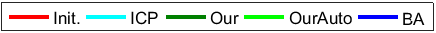}
	\caption{Error distribution of pose estimates for camera \textbf{\textit{C}2} in a two camera setup. 100 2D and 100 3D feature points are used. The following methods are compared: Init. - Initial pose obtained using a DLT like approach (2D feature points and corresponding 3D points are used)~\cite{Bougouet, Hartley2004}; ICP - refinement of Init. using Iterative Closest Point (only 3D feature points are used); Our - refinement of Init. using our bi-objective optimization with known $w$ (2D and 3D feature points are used); OurAuto - refinement of Init. using our bi-objective optimization with unknown $w$ (2D and 3D feature points are used); BA - refinement of Init. using Bundle adjustment (only 2D feature points are used). Gaussian noise is added to the data, being the variance of the 3D noise fixed ($\sigma_{3D}=18$mm), and the 2D noise $\sigma_{2D}$ is varied between $0.2$ and $1.8$ pixels (horizontal axes).}
	\label{fig_1} 
\end{figure*}
We test the performance of the proposed scheme under various conditions by varying the number of cameras and their positions as shown in \figurename~\ref{fig_C2} and \figurename~\ref{fig:C4.1}, by varying the noise magnitude in 2D and 3D feature points as explained above, and by varying the number of 2D and 3D feature points. For each configuration, 50 noise realizations are generated. For each noise realization, 2D feature points and their corresponding noisy 3D measurements from vertex maps are used to initialize the pose estimates via a Direct Linear Transform (DLT) based approach~\cite{Bougouet, Hartley2004}. Using the initial pose estimates, optimization is carried out via the proposed scheme, with known noise parameters as explained in Section~\ref{Proposed Approach}, and with unknown noise parameters using the automatic iterative estimation scheme as explained in Section~\ref{AutomaticWeighting} (required 3 iterations to converge in most cases).  Furthermore, optimization is also carried out via ICP algorithm using 3D feature points only, and via BA algorithm using 2D feature points only.
\begin{figure*}[t]
	\centering
	\subfigure[Rotation error]{\includegraphics[width=0.475\columnwidth]{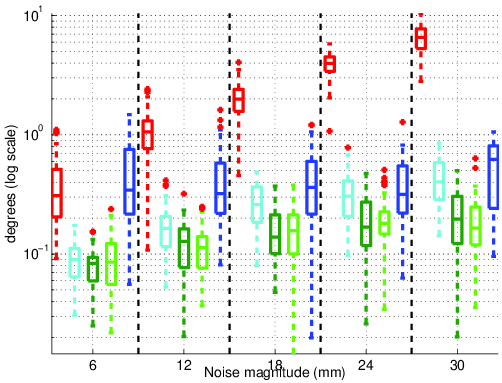}}
	\subfigure[Relative translation error]{\includegraphics[width=0.475\columnwidth]{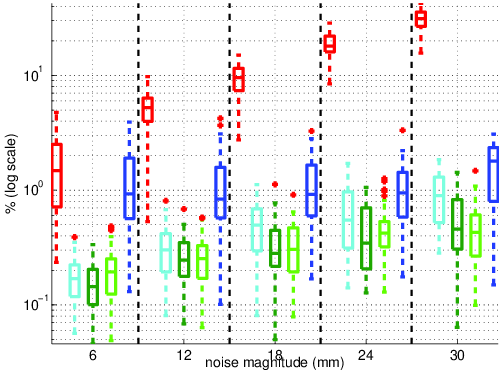}}
	\includegraphics[width=0.5\textwidth]{Figures/BAICPJounral/Legend2.png}
	\caption{Error distribution of pose estimates for camera \textbf{\textit{C}2} in a two camera setup. 100 2D and 100 3D feature points are used. Gaussian noise is added to the data, being the variance of the 2D noise fixed ($\sigma_{2D}=1$pix), and the 3D noise $\sigma_{3D}$ is varied between $8$mm and $30$mm (horizontal axes).}
	\label{fig_2} 
\end{figure*}

Accuracy of the estimated poses is computed by comparison with the ground truth poses as done in~\cite{Zhang04}. Two measures of accuracy are computed. First is the angular magnitude of residual rotation computed via $\hat{\mathbf{R}}^T_l\mathbf{R}_l$, and second is the relative translation error which is computed via $\frac{\|\hat{\mathbf{t}}_l-\mathbf{t}_{l}\|}{\|\mathbf{t}_l\|}$. The results of 50 realizations showing the accuracy, of each initialization and of each refinement approach, for each configuration are plotted by using the function \textit{boxplot} in MATLAB as shown in \figurename~\ref{fig_1}~-~\ref{fig_8}. The horizontal line inside each box marks the median, the edges mark the 25th and the 75th percentiles, the whisker edges show most extreme data points with outliers plotted separately as red crosses. 

The implementation of the proposed bi-objective optimization scheme and ICP is based on the non-linear optimization via Levenberg Marquardt (LM) algorithm~\cite{LM}, while the implementation of BA is based on a sparse variant of the LM algorithm called Sparse Bundle Adjustment (SBA)~\cite{sfmtoolbox, BA}.
\subsection{System Composed of Two Sensors}
\begin{figure*}[t]
	\centering
	\subfigure[Rotation error]{\includegraphics[width=0.475\columnwidth]{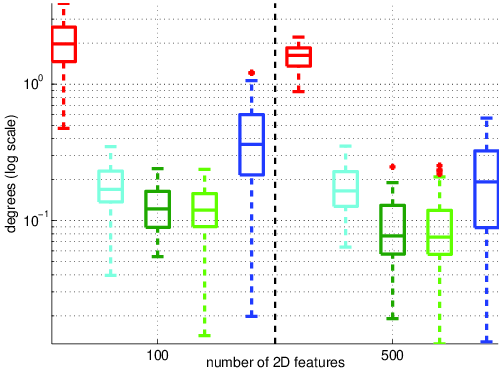}}
	\subfigure[Relative translation error]{\includegraphics[width=0.475\columnwidth]{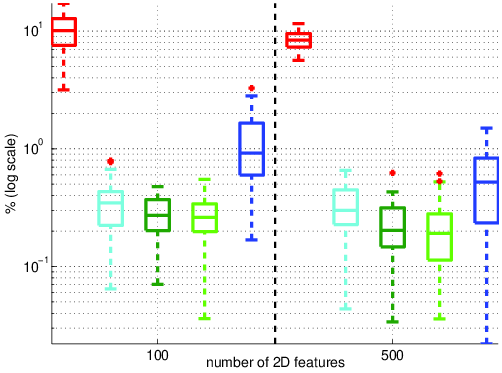}}
	\includegraphics[width=0.5\textwidth]{Figures/BAICPJounral/Legend2.png}
	\caption{Error distribution of pose estimates for camera \textbf{\textit{C}2} in a two camera setup. Gaussian noise is added to the data ($\sigma_{2D}=1$pix, $\sigma_{3D}=18$mm), $250$ 3D feature points and a varying number of 2D feature points (horizontal axes) is considered.}
	\label{fig_3} 
\end{figure*}

\begin{figure*}[h!]
	\centering
	\subfigure[Rotation error]{\includegraphics[width=0.475\columnwidth]{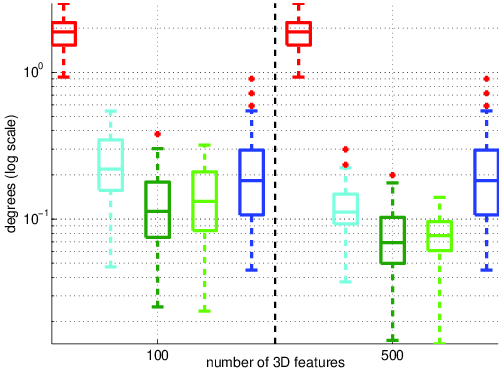}}
	\subfigure[Relative translation error]{\includegraphics[width=0.475\columnwidth]{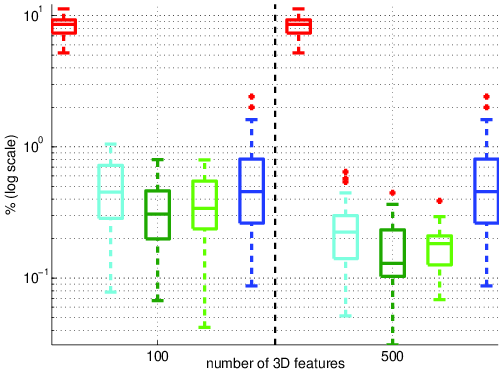}}
	\includegraphics[width=0.5\textwidth]{Figures/BAICPJounral/Legend2.png}
	\caption{Error distribution of pose estimates for camera \textbf{\textit{C}2} in a two camera setup. Gaussian noise is added to the data ($\sigma_{2D}=1$pix),$\sigma_{3D}=18$mm), $250$ 2D feature points and a varying number of 3D feature points (horizontal axes) is considered.}
	\label{fig_4} 
\end{figure*}

This section compares the performance of the proposed bi-objective optimization scheme, with known and unknown weighting factor, ICP and BA for refinement of camera pose parameters in a two camera setup shown in \figurename~\ref{fig_C2}. The pose of camera \textbf{\textit{C}2} with respect to camera \textbf{\textit{C}1} is estimated. After initialization, pose refinement is carried out using the four refinement methods and results are plotted in \figurename~\ref{fig_1}~-~\ref{fig_5}.
\subsubsection{Varying Noise Levels}
In this experiment, the extrinsic calibration is carried out using $100$ 2D feature points and $100$ 3D feature points. \figurename~\ref{fig_1} shows the error distribution for fixed 3D noise and varying 2D noise, while \figurename~\ref{fig_2} shows the distribution in case the 2D noise is kept fixed, and the 3D noise is varied.

As expected, the accuracy of the extrinsic calibration decreases with increasing noise levels. Also, all pose refinement approaches are able to improve the initial pose estimates, explained by the fact that only inlier data points are generated (no wrong matching feature points are included). A careful analysis of the results shows that our bi-objective optimization scheme with known $w$, which uses simultaneously the 2D and 3D data, provides better pose estimations when compared to ICP and BA, where only 3D feature points and 2D feature points are used, respectively. The weighting factor based on the noise variance information in \eqref{eq4.5} automatically gives prominence to more reliable data, decreasing the impact of the other sensor modality.  Moreover, it shows that our proposed automatic iterative estimation scheme used in the absence of information regarding noise parameters, and hence unknown $w$, is robust and also more accurate when compared to BA and ICP, and in most cases nearly as accurate as the method with known $w$.
\subsubsection{Varying Data Ratio}
In this experiment, the extrinsic calibration is carried out using fixed noise variance ($\sigma_{2D}=1$pix, $\sigma_{3D}=18$mm). \figurename~\ref{fig_3} shows the error distribution for a fixed number of 3D points and a varying number of 2D points, while \figurename~\ref{fig_4} shows the distribution in case the 2D points are kept fixed, and the number of 3D points is varied. Since the initial poses are obtained by using 2D feature points and their corresponding 3D points, the initialization varies in \figurename~\ref{fig_3} as number of 2D feature points vary but stays approximately the same in \figurename~\ref{fig_4} as the number of 2D feature points remain fixed. The conclusions drawn in the previous section regarding improved accuracy of the proposed approaches hold, and these results show that the proposed scheme generalizes for different ratios between the number of 2D and 3D points. Increasing the number of data points of one of the sensor modalities always improves the extrinsic calibration accuracy for the algorithms using those modalities. 
\begin{figure*}[t]
	\centering
	\subfigure[Mean rotation error]{\includegraphics[width=0.475\columnwidth]{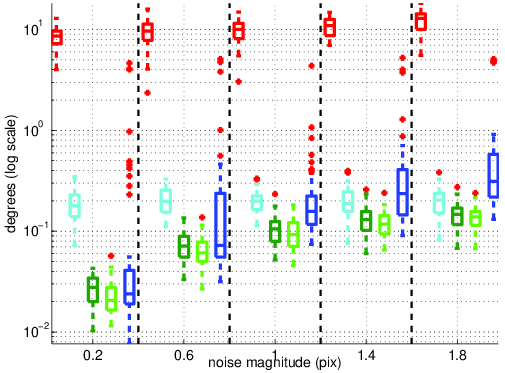}}
	\subfigure[Mean relative translation error]{\includegraphics[width=0.475\columnwidth]{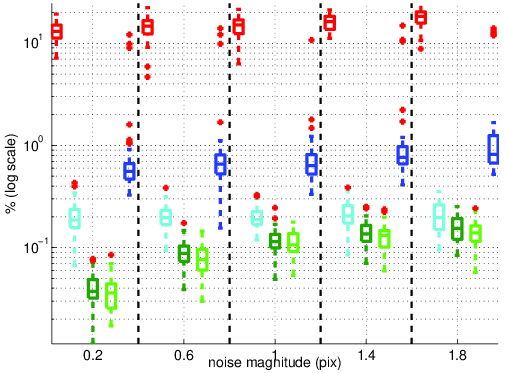}}
	\includegraphics[width=0.5\textwidth]{Figures/BAICPJounral/Legend2.png}
	\caption{Mean error distribution, of pose estimates for cameras \textbf{\textit{C}2}, \textbf{\textit{C}3} and \textbf{\textit{C}4}.  in a four camera setup. 100 2D and 100 3D feature points are used.  Gaussian noise is added to the data, being the variance of the 3D noise fixed ($\sigma_{3D}=18$mm), and the 2D noise $\sigma_{2D}$ is varied between $0.2$ and $1.8$ pixels (horizontal axes).}
	\label{fig_5} 
\end{figure*}
\begin{figure*}[!h]
	\centering
	\subfigure[Mean Rotation Error]{\includegraphics[width=0.475\columnwidth]{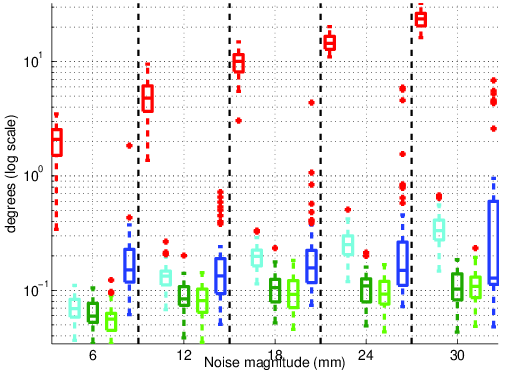}}
	\subfigure[Mean Relative Translation Error]{\includegraphics[width=0.475\columnwidth]{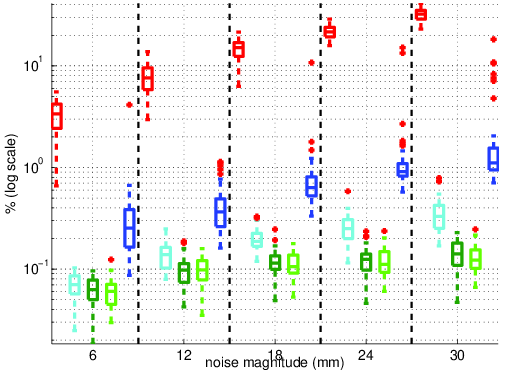}}
	\includegraphics[width=0.5\textwidth]{Figures/BAICPJounral/Legend2.png}
	\caption{Mean Error distribution, of pose estimates for cameras \textbf{\textit{C}2}, \textbf{\textit{C}3} and \textbf{\textit{C}4}.  in a four camera setup. 100 2D and 100 3D feature points are used. Gaussian noise is added to the data, being the variance of the 2D noise fixed ($\sigma_{2D}=1$pix), and the 3D noise $\sigma_{3D}$ is varied between $8$mm and $30$mm (horizontal axes).}
	\label{fig_6} 
\end{figure*}
Moreover, the results in \figurename~\ref{fig_1}, \figurename~\ref{fig_2}, \figurename~\ref{fig_3}, and \figurename~\ref{fig_4} show the increased robustness of the proposed approach and ICP to bad initialization as compared to BA.
\begin{figure*}[t]
	\centering
	\subfigure[Rotation Error]{\includegraphics[width=0.475\columnwidth]
		{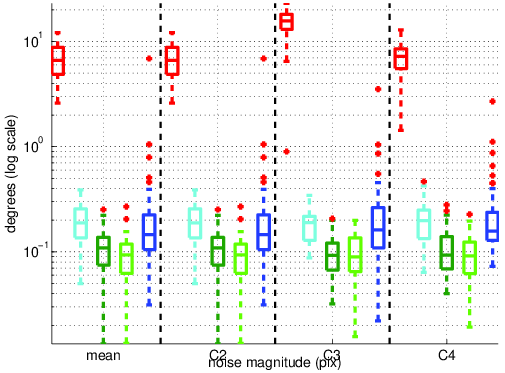}} 
	\subfigure[Relative Translation Error]{\includegraphics[width=0.475\columnwidth]{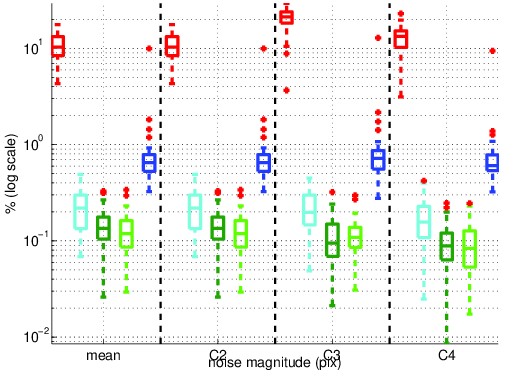}}
	\includegraphics[width=0.5\textwidth]{Figures/BAICPJounral/Legend2.png}
	\caption{Comparison of error distributions, of the extrinsic calibration of a four camera setup, using 100 2D and 100 3D feature points. The results are based on mean error distribution and error distribution for camera \textbf{\textit{C}2}, camera \textbf{\textit{C}3} and camera \textbf{\textit{C}4}. Gaussian noise is added to the data, being the variance of the both 2D noise and 3D noise fixed ($\sigma_{2D}=1$pix, $\sigma_{3D}=18$mm).}
	\label{fig_6.5} 
\end{figure*}
\begin{figure*}[!h]
	\centering
	\subfigure[Mean rotation error]{\includegraphics[width=0.475\columnwidth]{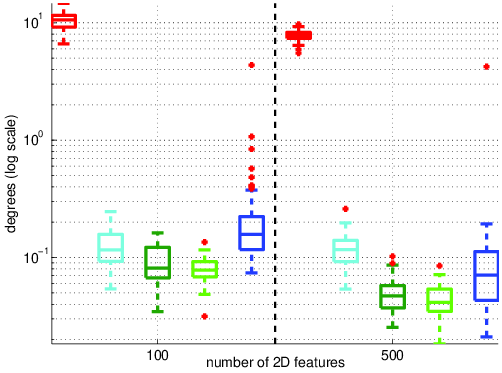}} 
	\subfigure[Mean relative translation error]{\includegraphics[width=0.475\columnwidth]{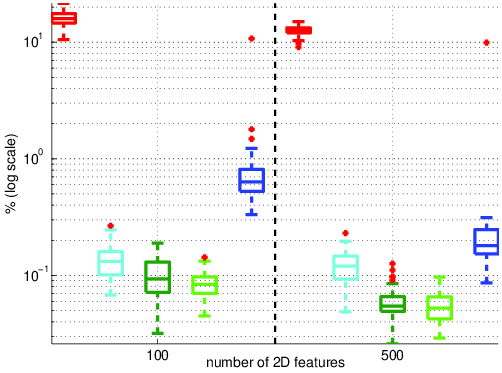}}
	\includegraphics[width=0.5\textwidth]{Figures/BAICPJounral/Legend2.png}
	\caption{Mean error distribution, of pose estimates for cameras \textbf{\textit{C}2}, \textbf{\textit{C}3} and \textbf{\textit{C}4}.  in a four camera setup. Gaussian noise is added to the data ($\sigma_{2D}=1$pix), $\sigma_{3D}=18$mm), $250$ 3D feature points and a varying number of 2D feature points (horizontal axes) is considered.}
	\label{fig_7} 
\end{figure*}
\begin{figure*}[thb]
	\centering
	\subfigure[Mean rotation error]{\includegraphics[width=0.475\columnwidth]{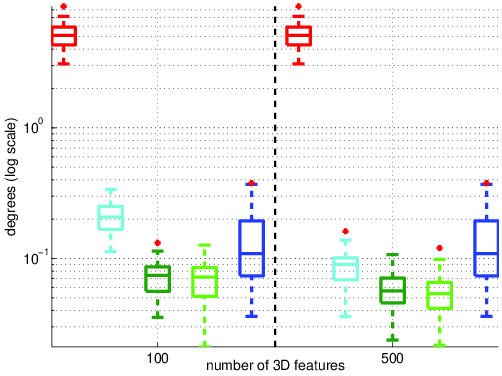}} 
	\subfigure[Mean relative translation Error]{\includegraphics[width=0.475\columnwidth]{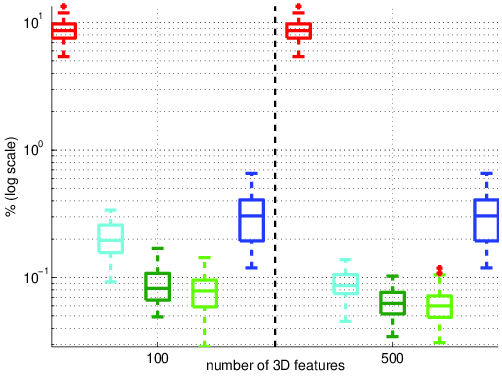}}
	\includegraphics[width=0.5\textwidth]{Figures/BAICPJounral/Legend2.png}
	\caption{Mean error distribution, of pose estimates for cameras \textbf{\textit{C}2}, \textbf{\textit{C}3} and \textbf{\textit{C}4}.  in a four camera setup. Gaussian noise is added to the data ($\sigma_{2D}=1$pix), $\sigma_{3D}=18$mm), $250$ 2D feature points and a varying number of 3D feature points (horizontal axes) is considered.}
	\label{fig_8} 
\end{figure*}
\subsection{System Composed of Four Sensors}
This section compares performance of the proposed bi-objective optimization scheme with ICP and BA for refinement of camera pose parameters in a four camera setup shown in \figurename~\ref{fig:C4.1}. The poses of cameras \textbf{\textit{C}2}, \textbf{\textit{C}3} and \textbf{\textit{C}4} are aligned with camera \textbf{\textit{C}1}. After initialization, pose refinement is carried out using the four refinement methods and results are plotted.
\subsubsection{Varying Noise Levels}
In this experiment, the extrinsic calibration is carried out using $100$ 2D feature points and $100$ 3D feature points. \figurename~\ref{fig_5} shows the mean error distribution for computed poses of all cameras, for fixed 3D noise and varying 2D noise. \figurename~\ref{fig_6} shows the mean distribution in case the 2D noise is fixed. These results again show the improved performance of the proposed approaches due to the use of both 2D and 3D information together, with the help of correct weighting factor. The performance of all methods gets affected as the noise in 2D and 3D data increases. These results also show improvement in performance of all methods as compared to the multi-view system composed of two cameras due to increased number of 2D and 3D points available. Moreover, these results show that the proposed scheme generalizes for different numbers of cameras used in the multi-view system.

We also notice an interesting behavior where in some cases the proposed automatic iterative scheme based on alternative computation of camera poses and $w$ gives better results compared to the scheme with known $w$. Apart from increase in the number of measurements per feature point, a reason for this can be that for the case of known $w$ we are assuming that for all the 2D and 3D feature points the variances of noise affecting them are the same and constant; but depending on a particular realization, the noise will be a bit higher or lower than the fixed value. Therefore the automatic procedure which tries to compute the variances directly from the noisy data is, in many cases, better able to capture the noise characteristics. For BA, \figurename~\ref{fig_6} shows a decrease in its performance as the 3D noise increases. The reason being that apart from its dependence on the initial camera poses, the initial guess of the 3D points corresponding to 2D feature points also gets worse due to increased 3D noise.

In \figurename~\ref{fig_6.5}, we compare the mean error distribution with error distributions of individual cameras for the single case of 2D and 3D noise variance  ($\sigma_{2D}=1$pix, $\sigma_{3D}=18$mm). These results show that while the initial guess for camera \textbf{\textit{C}3} is comparatively worse, the performance of optimization schemes is comparable across all views. 
\subsubsection{Varying Number of Points}
In this experiment, the extrinsic calibration is carried out using a fixed noise variance ($\sigma_{2D}=1$pix, $\sigma_{3D}=18$mm). \figurename~\ref{fig_7} shows the mean error distribution for a  fixed number of 3D points and a varying number of 2D points. \figurename~\ref{fig_8}, on the other hand, shows the mean distribution in case the 2D points are kept fixed, and the number of 3D points are varied. Here, again, the conclusions drawn in the previous sections hold, while also showing that increasing the number of data points of one of the sensor modalities always improves the extrinsic calibration accuracy for the methods using those modalities.
\section{Experiments with Real Data}\label{Experiments with real data}
\begin{figure*}[t]
	\centering
	{\includegraphics[width=0.7\columnwidth]{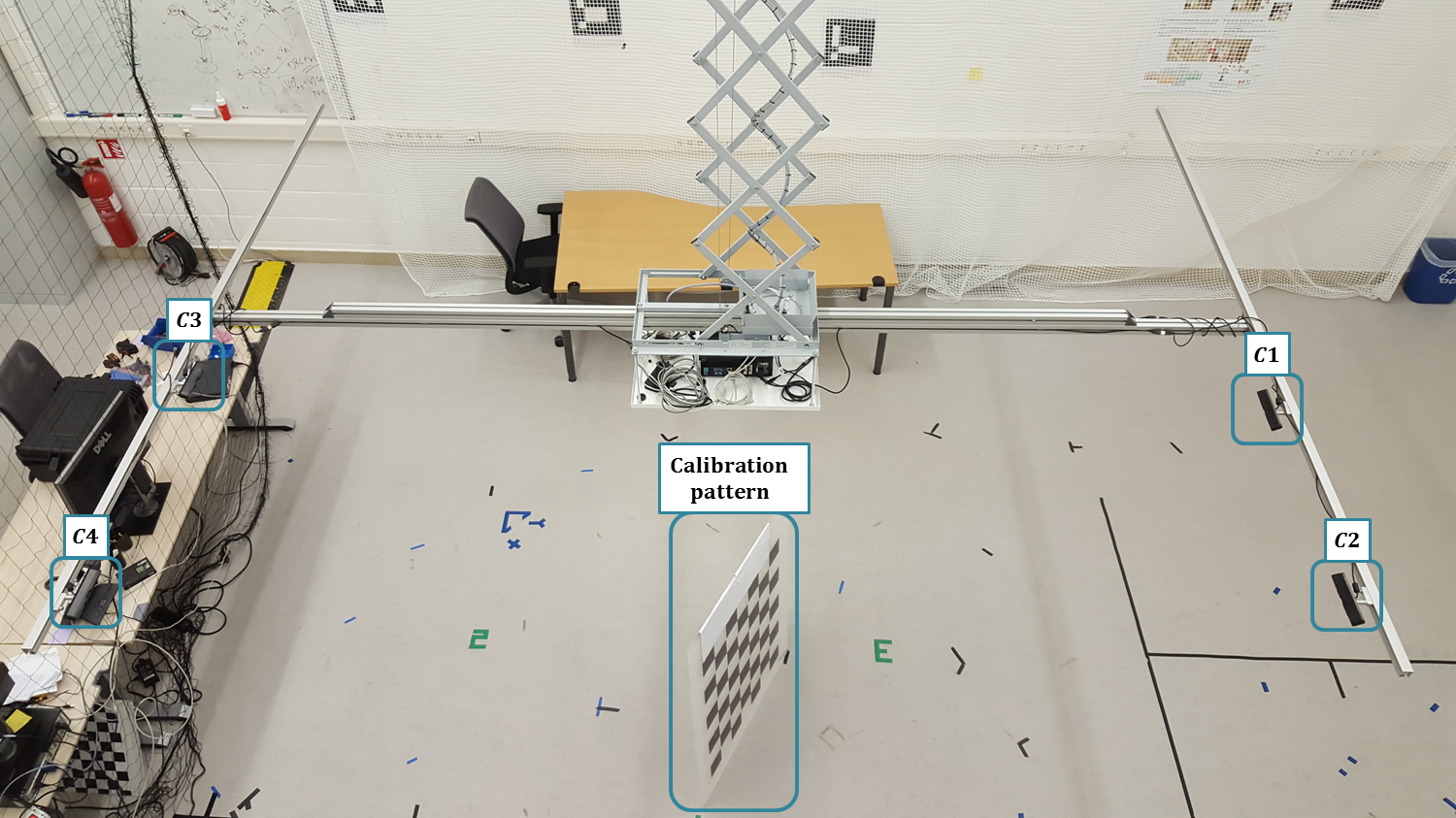}}
	\caption{Multi-view system consisting of 4 Asus Xtion Pro Live Cameras \textbf{\textit{C}1},  \textbf{\textit{C}2},  \textbf{\textit{C}3} and  \textbf{\textit{C}4} mounted on a ceiling lift. This system is used to acquire measurements of a real scene. A two-sided planar checkerboard calibration pattern used to extract feature points is also shown.}
	\label{fig_9} 
\end{figure*}

In this section, we carry out a qualitative performance analysis of the proposed bi-objective refinement scheme using a real setup. Our setup consists of 4 Asus Xtion Pro Live cameras~\cite{Xtion} with their positions shown in~\figurename~\ref{fig_9}. Each camera acquires an RGB image and a depth image which is mapped to the RGB image.
\begin{figure*}[h]
	\centering
	{\includegraphics[width=0.75\columnwidth]{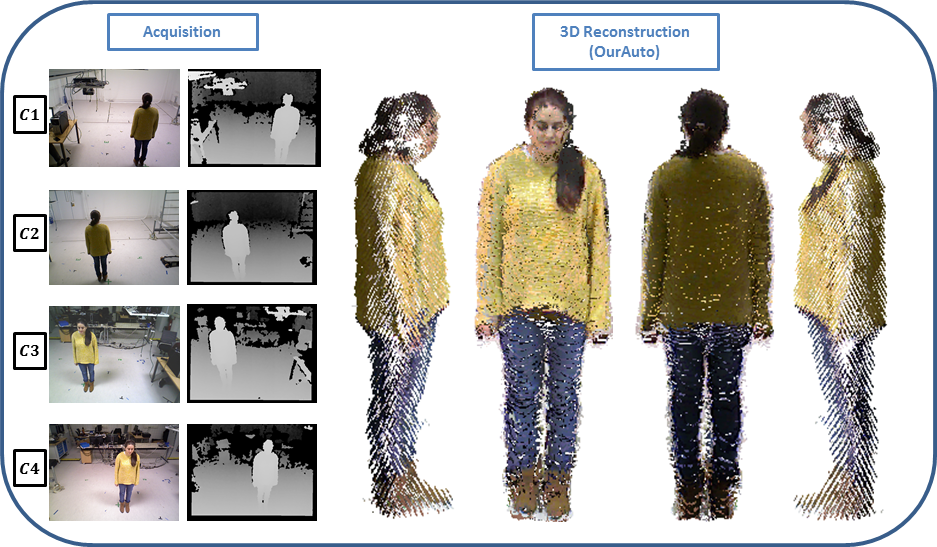}} 
	\caption{3D reconstruction of a human using a real scene acquired from the multi-view system shown in~\figurename~\ref{fig_9}. Acquisition: Each of the 4 cameras acquire an RGB image and a depth image. 3D Reconstruction: Point clouds based 3D reconstruction using pose estimates refined by the proposed bi-objective scheme with the help of automated weighting.}
	\label{fig_10} 
\end{figure*}
The first step is to perform intrinsic calibration to find the intrinsic and distortion parameters for each camera. For this purpose, we use the method proposed by Zhang~\cite{Zhang99} which uses 2D corners extracted from RGB images of a checkerboard pattern viewed at different poses to compute these parameters~\cite{Bougouet}. As mentioned before, the measurements of these RGB-D cameras suffer from inherent depth bias. Therefore, we perform a depth bias correction procedure, similar to the one used in~\cite{Nakazawa12}, for each camera separately. This procedure requires placing the camera at known distances away from an object (a plane in our case). Using known and measured depth values, we estimate the coefficients of a polynomial which computes the depth correction as a function of measured depth value. These coefficients are unique to each camera and, hence, are used to correct the depth measurements acquired by that camera.

\begin{figure*}[t]
	\centering
	{\includegraphics[width=0.75\columnwidth]{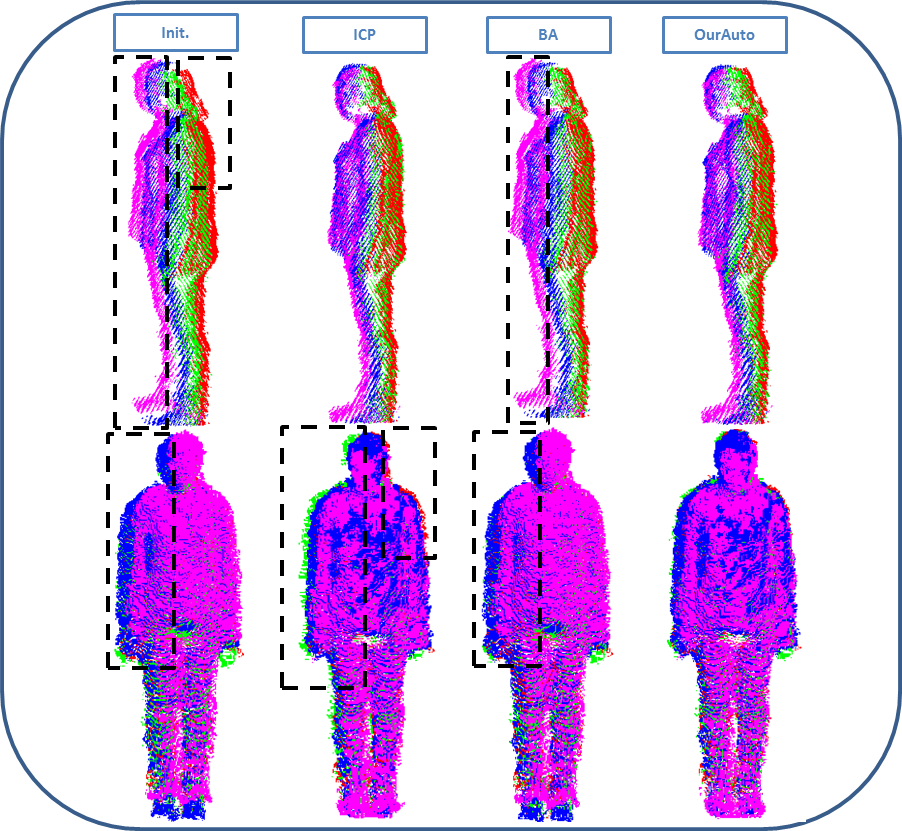}} 
	\caption{Comparison of 3D reconstructions of a human using a real scene as shown in~\figurename~\ref{fig_9}, via different calibration methods namely Init., ICP, BA and  OurAuto described in~\figurename~\ref{fig_2}. The acquisitions from cameras \textbf{\textit{C}1}, \textbf{\textit{C}2}, \textbf{\textit{C}3} and \textbf{\textit{C}4} are assigned the colors red, green, blue and magenta, respectively. Misalignments are highlighted via black boxes. \textit{Top Row} shows side view of the 3D reconstruction and misalignment of views in the results of Init. and BA can be seen clearly, while \textit{Bottom Row} shows the frontal view and misalignment of views in the results of Init., ICP and BA are visible. It can also be seen that OurAuto gives better results compared to the other methods.}
	\label{fig_11} 
\end{figure*}

After intrinsic calibration and depth bias correction, the next step is to perform the extrinsic calibration using the proposed bi-objective scheme. We first need to extract matching 2D and 3D feature points using RGB and depth images acquired by all 4 cameras. We again use different views of a (two-sided) planar checkerboard pattern as shown in~\figurename~\ref{fig_9} and extract matching corners from RGB images to be used as 2D feature points and use the corresponding depth values from depth images to get the 3D feature points. The 3D feature points are filtered via a plane detection approach based on {RANSAC} algorithm to remove outliers if any exist. For this experiment, only 59 2D and 3D feature points were used. The initial pose estimates are generated in the same manner as explained in Section~\ref{Experiments with synthetic data}, via a Direct Linear Transform (DLT) based approach~\cite{Bougouet, Hartley2004}. These initial poses are then refined via the proposed iterative pose estimation and weight estimation approach explained in Section~\ref{AutomaticWeighting}, BA and ICP. Once the refined poses are obtained, they can be used to produce full, textured, 3D reconstructions using data acquired by all 4 cameras as shown in~\figurename~\ref{fig_10}. A qualitative comparison of 3D reconstructions obtained via different calibration methods is shown in~\figurename~\ref{fig_11}. It can be seen that the partial reconstructions are better aligned using the proposed method, which means that the quality of the extrinsic calibration is superior when compared to the other approaches. Note that we are only showing the alignment of the partial point clouds, and no post-processing step such as smoothing or meshing are applied. We chose to do so to better assess, visually, the accuracy of the extrinsic calibration.

\section{Conclusion}\label{Conclusion}
In this paper we have proposed a framework for RGB and depth sensor fusion based on bi-objective optimization, for refinement of extrinsic calibration in RGB-D multi-view systems. Our bi-objective optimization scheme makes use of a cost function from the BA algorithm for 2D feature points extracted from RGB images and a cost function from the ICP algorithm for 3D feature points extracted from depth images. We analytically derive an expression for the weighted bi-objective cost function. It also analytically relates the weighing factor to the noise in the 2D and 3D measurements, thus making the cost function free of any parameter that needs to be tuned. In case the information regarding measurement noise in 2D and 3D data is not available, we propose an iterative scheme which alternates between estimation of noise parameters assuming known poses, and estimation of camera poses assuming known noise parameters. Thus, it enables us to automatically compute the correct weighting factor when information about measurement noise is not available. A thorough investigation of the performance of the proposed approach for both synthetic and real data showed improved accuracy compared to refinement schemes which only use 2D or 3D information, and comparative performance of proposed approaches with known and unknown noise parameters. These experiments also showed the invariance of the proposed approach under various conditions which include varying the number and position of cameras, varying the 2D and 3D noise and varying the number of the 2D and 3D feature points.
\section{Acknowledgment}
This work was supported by the National Research Fund (FNR), Luxembourg, under the CORE project C11/BM/1204105/FAVE/Ottersten.

\section{Appendix}\label{Appendix}
\subsection{Non-linear Optimization for Proposed Bi-Objective Framework}\label{NumericalEstimation}
Due to the non-linear dependence of cost function in~\eqref{eq4.5} on parameters in $\mathbf{S}$, the MLE $\hat{\mathbf{S}}$ is to be computed via a numerical scheme based on non-linear optimization. In this scheme at every iteration a small change is introduced in the current set of parameters leading to comparatively improved performance or lower residual~\cite{MLE}.
First step in this scheme is to linearize $\mathbf{b}_{l,m}^{j}({\mathbf{T}}_{l},{\mathbf{T}}_{m})$ and $\mathbf{a}_{l}^{h}({\mathbf{S}}_{l}^{h})$ about current estimate $\hat{\mathbf{S}}$ assuming very small error $\Delta{\mathbf{S}}$ using Taylor expansion to get:
\begin{align}\label{eqA5}
\mathbf{b}_{l,m}^{j}({\mathbf{T}}_{l},{\mathbf{T}}_{m}) \approx \mathbf{b}_{l,m}^{j}(\hat{\mathbf{T}}_{l},\hat{\mathbf{T}}_{m}) + \mathbf{J}_{\mathbf{b}_{l,m}^{j}}\Delta{\mathbf{S}},
\end{align}
and:
\begin{align}\label{eqA6}
\mathbf{a}_{l}^{h}({\mathbf{S}}_{l}^{h}) \approx \mathbf{a}_{l}^{h}(\hat{\mathbf{S}}_{l}^{h}) + \mathbf{J}_{\mathbf{a}_{l}^{h}}\Delta{\mathbf{S}},
\end{align}
where $\mathbf{J}_{\mathbf{b}_{l,m}^{j}}$ and $\mathbf{J}_{\mathbf{a}_{l}^{h}}$ are Jacobians of $\textbf{b}_{l,m}^{j}({\mathbf{T}}_{l},{\mathbf{T}}_{m})$ and  $\mathbf{a}_{l}^{h}({\mathbf{S}}_{l}^{h})$, with respect to ${\mathbf{S}}$, respectively.
Replacing \eqref{eqA5} and \eqref{eqA6} in \eqref{eq4.5} and concatenating $\mathbf{b}_{l,m}^{j}$, $\mathbf{a}_{l}^{h}$, $\Delta{\mathbf{S}}$ and corresponding Jacobians we have:
\begin{align}\label{eqA7}
V({\mathbf{S}}) &\approx  (\mathbf{B}+\mathbf{J}_\mathbf{B}\mathbf{\Delta}{\mathbf{S}})^T  (\mathbf{B}+\mathbf{J}_\mathbf{B}\mathbf{\Delta}{\mathbf{S}}) + \nonumber \\
&\quad  w(\mathbf{A}+\mathbf{J}_\mathbf{A}\mathbf{\Delta}{\mathbf{S}})^T
(\mathbf{A}+\mathbf{J}_\mathbf{A}\mathbf{\Delta}{\mathbf{S}}) \nonumber \\
&=(\mathbf{B}^T\mathbf{B}+2{\Delta}{\mathbf{S}}^T\mathbf{J}_\mathbf{B}^T
\mathbf{B}+{\Delta}{\mathbf{S}}^T\mathbf{J}_\mathbf{B}^T
\mathbf{J}_\mathbf{B}{\Delta}{\mathbf{S}}) \nonumber \\
& \quad + w(\mathbf{A}^T\mathbf{A}+2{\Delta}{\mathbf{S}}^T\mathbf{J}_\mathbf{A}^T
\mathbf{A}+{\Delta}{\mathbf{S}}^T\mathbf{J}_\mathbf{A}^T
\mathbf{J}_\mathbf{A}{\Delta}{\mathbf{S}}).
\end{align}
After that we take the derivative of $V({\mathbf{S}})$ with respect to ${\mathbf{S}}$ and equate it to zero to get:
\begin{align}\label{eqA8}
\frac{\partial{V({\mathbf{S}})}}{\partial{{\mathbf{S}}}} \approx & \mathbf{J}_\mathbf{B}^T\mathbf{B} + \mathbf{J}_\mathbf{B}^T
\mathbf{J}_\mathbf{B}\mathbf{\Delta}{\mathbf{S}}
+ w\mathbf{J}_\mathbf{A}^T\mathbf{A} + w\mathbf{J}_\mathbf{A}^T
\mathbf{J}_\mathbf{A}\mathbf{\Delta}{\mathbf{S}}=0.
\end{align}
Rearranging \eqref{eqA8}, we get the parameter update rule as:
\begin{align}\label{eqA9}
\mathbf{\Delta}{{\mathbf{S}}}  =   -(&\mathbf{J}_\mathbf{B}^T
\mathbf{J}_\mathbf{B} + w\mathbf{J}_\mathbf{A}^T
\mathbf{J}_\mathbf{A})^{-1}
(\mathbf{J}_\mathbf{B}^T\mathbf{B} + w\mathbf{J}_\mathbf{A}^T\mathbf{A}).
\end{align}
We can also rearrange \eqref{eqA9} according to Levenberg-Marquardt LM~\cite{LM} algorithm get the parameter update rule as:
\begin{align}\label{eqA11}
((\frac{1}{2\sigma_{3D}^{2}}\mathbf{J}_\mathbf{B}^T
&\mathbf{J}_\mathbf{B} + \frac{1}{\sigma_{2D}^{2}}\mathbf{J}_\mathbf{A}^T
\mathbf{J}_\mathbf{A}) + \nonumber \\
&\lambda \mathtt{diag}(\frac{1}{2\sigma_{3D}^{2}}\mathbf{J}_\mathbf{B}^T
\mathbf{J}_\mathbf{B} + \frac{1}{\sigma_{2D}^{2}}\mathbf{J}_\mathbf{A}^T
\mathbf{J}_\mathbf{A}))\mathbf{\Delta}{\mathbf{S}} \nonumber \\
&= -(\frac{1}{2\sigma_{3D}^{2}}\mathbf{J}_\mathbf{B}^T\mathbf{B} + \frac{1}{\sigma_{2D}^{2}}\mathbf{J}_\mathbf{A}^T\mathbf{A}),
\end{align}
where $\lambda$ is the damping factor.

\end{document}